\documentclass[11pt]{article}
\usepackage{mtsummit2019}
\usepackage{times}
\usepackage{url}
\usepackage{latexsym}
\usepackage[small,bf]{caption} 
\usepackage[utf8]{inputenc}

\title{Improving Robustness in Real-World Neural Machine Translation Engines}

\date{}
\author{Rohit Gupta, Patrik Lambert, Raj Nath Patel and John Tinsley\thanks{All authors contributed equally}\\
  Iconic Translation Machines Ltd.\\
  Invent Building, DCU Campus, Glasnevin,\\
  Dublin 9, Ireland.\\
  {\tt rohit,patrik,raj,john@iconictranslation.com}  }
\date{}
\begin{document}
\maketitle
\begin{abstract}
As a commercial provider of machine translation, we are constantly training engines for a variety of uses, languages, and content types. In each case, there can be many variables, such as the amount of training data available, and the quality requirements of the end user. These variables can have an impact on the robustness of Neural MT engines. On the whole, Neural MT cures many ills of other MT paradigms, but at the same time it has introduced a new set of challenges to address. In this paper, we describe some of the specific issues with practical NMT and the approaches we take to improve model robustness in real world scenarios.
\end{abstract}

\section{Introduction}

As a commercial provider of bespoke machine translation (MT) solutions for enterprise users, we train engines all day, every day for a variety of different languages, and content types, with different quantities and quality of training data. On a case by case basis, there are a lot of variables to contend with. 

The breakthrough of Neural MT (NMT) over the past number of years, and the step change in quality it can produce, means that it is a no-brainer to adopt and make an integral part of our technology stack. However, there are still some practical gaps that need to be addressed in the core technology, in order to make it broadly production ready and flexible. These are either specific issues or topics that were already resolved in Statistical MT and have been reintroduced, or new types of issues unique to neural models.

This can include, but is not limited to, the need for a  more rigorous data cleaning step, a lack of robustness around handling terminology and various types of mistranslations, and the ability to adapt to different domains.

Sometimes we can handle these issues elegantly in the models, but certain variables such as the volume of training data available in each case, make it a little less predictable. In some cases,  we have to find more practical workarounds in our data preparation, and pre- and post-processing steps, in order to get engines production ready.

In this paper, after giving an overview of our NMT pipeline, we will focus on how we address the following issues in order to better prepare NMT engines for real-world deployment: 1) data cleaning, 2) over-generation, 3) improving robustness when translating entities, and 4) domain adaptation.

\subsection{Our Pipeline}
Our NMT pipeline is composed of several components, which are described in the following sections. Training data is first processed through a corpus preparation pipeline. This pipeline includes data cleaning and filtering scripts (see Section \ref{sect:corprep}), as well as a processing pipeline. At test time, this processing pipeline is applied to the source text. Do-not-translate words are replaced by placeholders and replaced back in the translation (Section \ref{sect:dnts}). This technique can also be used to force the translation of specific terminology. Before training, the tokens are split into sub-words to limit the vocabulary size (see Section~\ref{sect:subwords}). The model is dynamically adapted to the source sentence if a similar segment is found in the training corpus (Section \ref{sect:adapt}). After translation, a post-processing module deletes over-generation patterns based on the source sentence (Section \ref{sec:cleaning:experiments}).

\section{Training Data cleaning}
\label{sect:corprep}
Garbage in, garbage out. This is more relevant than ever for NMT which has been shown to be more sensitive to noisy data. In the following, we describe some steps we take to prepare different corpora prior to training.
\subsection{Description}
The data cleaning pipelines includes the following steps:
\begin{itemize}
    \item \textbf{Character and encoding cleaning:} cleans encoding issues, cleans and normalizes incorrect characters.
    \item \textbf{Punctuation and digit filtering:} the following sentence pairs are filtered: (i) if in one of the sides, less than half of the characters are digits or letters; and (2) if in one of the sides, the sentences is only composed of digits and spaces. The intuition behind these steps is that sentence pairs formed mostly by punctuation or digit are not very useful for training and thus can be discarded.
    \item \textbf{Copy filtering:} sentence pairs in which the target side is a copy of the source side are filtered out. It has been observed that copied sentences are very harmful for NMT~\cite{khayrallah-koehn-2018-impact}.
    \item \textbf{Duplicate removal:} in this step, repeated sentence pairs are removed.
    \item \textbf{Length-based filter:} sentence pairs in which one of the sides has less characters than a threshold are filtered out, as well as sentence pairs in which the length ratio is less than a threshold. Specifically, the average ratio of source and target sentence length in the training corpus is first calculated, as well as its standard deviation. The sentence pairs whose ratio differ more than 6 standard deviations from the average are discarded.
    \item \textbf{Language-based filter:} sentence pairs whose respective language are not the correct one are discarded. The language identification is performed in two stages. First, the main script of the sentence is identified. Based on this information, the set of possible languages is determined. If the correct language is not part of the set of possible languages, the sentence is discarded. Otherwise, the language identification is performed within the set of possible languages. To limit the number of false negatives, we split the sentence in two and consider that the language is incorrect only if both halves have been classified as the same incorrect language.
    \item \textbf{Do-not-translate word replacement:} words and phrases detected as do-not-translated entities are replaced by a placeholder if they appear in both sides of the sentence pair.
    \item \textbf{Processing pipeline:} each side of the training corpus is processed independently with processors pertinents for the task at hand, including tokenization and truecasing.
    \item \textbf{training/development/test sets splitting:} the splitting strategy ensures the same distribution of sentences with do-not-translate entities as well as of each length range in the development and test data. It also keeps 5\% of development set sentences overlapping with the training set, which is helpful for training.
\end{itemize}

  \begin{table}[!hbt]
	\centering
	\begin{tabular}{l|c|c}
		  & sentence pairs & English words \\ \hline
		 Train (Iconic) & 202,249 & 1,868,403 \\ \hline
		 Train (Moses)& 205,434 & 1,884,124 \\ \hline
		 Dev (Iconic) & 2000 & 22,502 \\ \hline
		 Dev (random) & 2000 & 19,501 \\ \hline
		 Test & 2100 & 24,571 \\
	\end{tabular}
	\caption{Statistics of KDE4 data for the training, development and test corpora processed by Iconic pipeline and Moses tools.}
	\label{tab:kde4:stats}
\end{table}

\begin{table*}[!hbt]
	\centering
	\begin{tabular}{l|c|c|c|c|c|c}
		 Corpus Preparation & BLEU & 1-TER & OVER & REP & UNDER & DROP\\ \hline
		 Moses tools & 31.4 $\pm$0.3& 47.2 $\pm$0.3& 29.3 $\pm$2.3& 5.1 $\pm$0.5& 8.0 $\pm$1.0& 9.5 $\pm$0.8\\ \hline
		 Iconic     & 33.7 $\pm$0.4& 50.2 $\pm$0.1& 29.3 $\pm$5.8& 3.3 $\pm$0.3& 8.3 $\pm$0.6& 8.7 $\pm$0.3\\ \hline
		 Iconic+DNT & 32.5 $\pm$0.2& 48.5 $\pm$0.3& 26.0 $\pm$6.0& 2.9 $\pm$0.1& 6.7 $\pm$1.2& 8.8 $\pm$0.5\\ \hline
		 Iconic + rep-del & 33.7 $\pm$0.3& 50.2 $\pm$0.1& 17.3 $\pm$2.1& 3.1 $\pm$0.2& 8.3 $\pm$0.6& 8.7 $\pm$0.2\\ \hline
		 Iconic+DNT+rep-del & 32.4 $\pm$0.2& 48.4 $\pm$0.3& 13.3 $\pm$3.1& 2.7 $\pm$0.1& 6.7 $\pm$1.2& 8.8 $\pm$0.5\\ \hline
		 
	\end{tabular}
	\caption{Evaluation scores for training on data processed by Moses tools, our pipeline without (Iconic) and with (Iconic+DNT) replacement of do-not-translate phrases by placeholders.}
	\label{tab:cleaning:results}
\end{table*}

 \subsection{Experiments}\label{sec:cleaning:experiments}
 We evaluated the impact of our data cleaning pipeline on the KDE4 German-English data, obtained from the OPUS corpus\footnote{http://opus.nlpl.eu/}. We compared the training with data processed by our pipeline and with data processed by Moses tools (tokenization, length-based filter and true-casing). We used the same length parameters for the length-based filter (175 words) and the same true-casing models. The statistics of the data are shown in Table~\ref{tab:kde4:stats}. In the case of the Moses pipeline, the development set was selected at random. The test set was the same, but processed according to each pipeline.

 We trained small transformer models with the Fairseq tool~\cite{ott-etal-2018-scaling}, with the same parameters as those indicated in the fairseq github site for IWSLT'14 German to English. We averaged the 5 checkpoints around the best model. We repeated the training 3 times and report the average and standard deviation of the 3 runs.
 
Results are reported in Table~\ref{tab:cleaning:results}. Training with our pipeline improves BLEU~\cite{papineni:2002} and TER~\cite{snover:2006:ter} scores respectively by 2.3 and 3.0 points. The difference is larger than standard deviation error bars, thus it is statistically significant according to this criterion. This suggests that efforts to better clean the data and to choose the validation set carefully are beneficial in terms of automated quality metrics. 

NMT models are not perfect at controlling the output length and sometimes drop or duplicate content. To evaluate this category of errors, the rest of metrics measure over-generation (repetitions) and under-generation (source text not covered). OVER simply counts repetitions in the output, while UNDER counts under-generation based on the ratio of number of source and output words. REP and DROP count respectively the number of repetitions and under-generation in the output based on the alignment with the source~\cite{malaviya-etal-2018-sparse}. Interestingly, the REP score is significantly lower with our pipeline. The DROP score average is also lower although the difference lies within the standard deviation. This suggests that the engine is more robust to under- and over-generation with our pipeline.

Replacing do-not-translate phrases (DNTs) by placeholders (see section~\ref{sect:dnts}) yields slightly worse BLEU and TER scores. However, the OVER, REP and UNDER scores are improved. Thus using DNTs may improve robustness. The worsening of BLEU and TER may be due to the fact that we used only one type of placeholder to replace entities which appear in different contexts (for example, URLs and numbers). Using different types would improve the modelling of each one.

 Our pipeline includes a module to detect duplicated content in the translation and to delete it. The detection is based on the source sentence. That is, if the source text contains a repetition, it is not incorrect to have it in the translation. To decide whether a repetition should be deleted or not, we adopted a conservative criterion favoring precision rather than recall. We delete repeated words if they are aligned with the same source word. The alignment may be given by the attention weights or by an external alignment. 
 
 Table~\ref{tab:cleaning:results} also shows the impact of using our source-based repetition deletion module ("rep-del"). This module drops the average number of repetitions (OVER) of the Iconic system from 29.3 to 17.3 and the REP score from 3.3 to 3.1. Applied to the system with DNTs, OVER drops from 26.0 to 13.3 and the REP score from 2.9 to 2.7. Thus this module is effective at removing repetitions, with no significant impact on BLEU and TER.



\section{Tokenization \& Subword Encodings}
\label{sect:subwords}

Whereas Statistical was fairly predictable in terms of how it would perform on certain inputs - for good or for bad - neural models can react in a peculiar manner on unseen input. This can manifest itself more with things like named entities and, often, these are of critical importance in real-world scenarios where they may refer to drug names, email address, defendent namess, etc., so the MT needs to be robust and predictable.

\subsection{Preparation}

After cleaning we tokenize and normalize our data. We also apply subword encodings. They are particularly helpful to limit the vocab size for an NMT system. Subwords also help in tackling out-of-vocabulary (OOV) problem in NMT. It helps in improving the coverage by splitting words. Therefore, the system can translate different forms of a word even if it was not seen during training. 

\subsection{Issues with Tokenization}
Too much tokenization can also cause issues. We often come across words and phrases which should be left untouched during translation. They are in general entities and they can represent file numbers, file paths, formatting tags, commands, product names, email address, URLs, terms etc. In Neural MT, this process of copying is also learned during translation ~\cite{knowles2018context}. However,  if we do not pay attention to such entities it gets difficult to recover then successfully as some parts of the entities may get modified during translation.  


Therefore, we focus on learning the translation part and normalize the other data where we require untouched copy as a part of pre-processing and post-processing. 

\subsubsection{Do Not Translate Terms}
\label{sect:dnts}
We define do-not-translate terms (DNTs) as terms which are exact copy from the source. They are neither translated nor transliterated. The languages where the source and target have different scripts and do not share characters, it is easier to determine such terms. For example, when translating from Chinese to English it is easy to spot English text in the Chinese sentence and such words are almost always exact copy from the source. The languages who share alphabets e.g. if both languages belong to Latin, in such language pairs, we need much context to determine. 

We determine following expressions as DNT terms:
\begin{itemize}
    \item Email addresses, URLs
    \item Numbers with two or more digits (without comma and dot)
    \item Any combination of number (at least two digits) and English characters
    \item File names and paths with valid extensions
    \item XML Tags
    \item English characters when the source is Non-Latin and target is English
\end{itemize}
\subsubsection{How DNTs are helpful?}
We detect DNTs in the source and replace them with a placeholder token during translation. For example, the following segment from the MultiUN dataset can be converted to have two DNTs (DNTID1 and DNTID2).

\begin{itemize}
\item "For more information about the project and all \textbf{19} targets, visit \textbf{www.post2015consensus.com}"
\item "For more information about the project and all \textbf{DNTID1} targets, visit \textbf{DNTID2}"
\end{itemize}

Here DNTID1 is 19 and DNTID2 is www.post2015consensus.com. The system learns to copy DNTID1 and DNTID2 placeholders instead of actual numbers and URLs. We issue multiple DNTs (here ID1 and ID2) so as to have position information when there are more than one DNTs in a sentence.

\subsubsection{Issues with Subwords}
\label{subsubsect:issues-subwords}
Subword translation is an approach used in NMT to tackle out-of-vocabulary (OOV) problem using byte-pair encoding (BPE) or other similar segmentation techniques. It is now defacto to use subwords in NMT as with the better vocab coverage it enables the NMT models with excellent copying capability. The copying behaviour is required when the named entities need to be copied from the source text to the target translation. Although subword NMT works quite well at copying, it sometimes fails to copy the complete sequence of subwords in the translated text and results in spelling errors. 

\subsubsection{Spelling Errors in Subword NMT}
\label{subsubsect:spelling-err}
In general, NMT models perform quite poorly on rare words,~\cite{luong2015stanford,sennrich2016neural,arthur2016incorporating} due to the fixed vocabulary of NMT models. The most common categories of rare words are named entities and nouns. These entities often pass through the NMT system unchanged. For example, the word "\textit{Gonzalez}" is broken into  "\textit{G@@ on@@ z@@ al@@ e@@ z}" by BPE and passes through the NMT system unchanged. However, when it fails, the model can drop or wrongly translate subwords which results in perceived misspellings.

\paragraph{Subword Dropped}\label{para:subword-dropped} In this case when a subword (which is part of a named entity) is not copied in the translated text. For example, the word \textit{Stephen} is split into "\textit{Ste@@ p@@ hen"} and say in the translation process NMT system failed to copy subword "p@@", then the resulting translation would be \textit{Stehen}.

\paragraph{Subword Translated}\label{para:subword-translated} In this case one or more subwords, which were meant to be copied, are actually translated. For example, in our German-English NMT system, the named entity \textit{littlebits} is translated as \textit{littlement}. It was due to the fact that applying byte pair encoding, the word \textit{littlebits} is split into "li@@ tt@@ leb@@ it@@ s" where the subwords \textit{@@it s} are translated as “\textit{ment}”. 

\subsection{Tackling Subword Issues}
\label{subbsec:tackling-subword}
We suggest that the above mentioned issues causing spelling errors in named entities are mainly because of over splitting. In BPE, the algorithm checks each subword in the given vocab and if not found, it will recursively split the segment into smaller units (by reversing byte-pair encoding merge operations) until all units are either in-vocabulary, or cannot be split further (often character level splits). For named entities, it is quite common to have unseen subwords resulting into character level splits. We propose two methods to resolve byte-pair encoding issues.
\paragraph{No More Split}
\label{para:no-more-split}

In this method, we restrict the encoding algorithm from splitting unseen subwords into characters. The intuition behind is that copying single unseen token would be easier than copying a sequence of characters.

\paragraph{Protect Unseen Words}
\label{para:protect-unseen}
After applying the encoding, this method counts the unseen subwords (not in-vocabulary) and if the count is more than a threshold value it keeps the original word. The logic is to use UNK-token translation transferring these entities in the target text. However, this method is highly dependent on accuracy of alignment and UNK-translation.

\begin{table}[!hbt]
	\centering
	\begin{tabular}{l|c|c|c}
		&  & \#segments & \#words \\ \hline
		& train & 160239 & 3998597 \\
		& dev & 7283 & 181021 \\
		& test & 6750 & 153697
	\end{tabular}
	\caption{Data distribution after cleaning and applying tokenizer (source side)}
	\label{tab:data}
\end{table}

\subsubsection{Experiments \& Results}
\label{subsubsec:exp-subword}
Our sample results here are based on the publicly available IWSLT dataset~\footnote{\url{https://wit3.fbk.eu/archive/2014-01/texts/de/en/de-en.tgz}}. The distribution of train, dev, and test datasets is detailed in Table~\ref{tab:data}. We randomly select a development set from the training data. The test set is created by combining dev (2010, 11), and test (2010, 11, 12) sets of earlier IWSLT shared tasks. 

We use a shared vocabulary BPE Model~\cite{sennrich2016neural} for subword segmentation, with a code of 32000 merge operations. We use convolutional~\cite{gehring2017convolutional} encoder-decoder (15x15) architecture with the size of hidden units and  word embedding of 512. For the training of model parameters, we use NAG ~\cite{qu2017accelerated} with cross entropy as a loss function. We start with a learning rate of 0.25 and reduce it by a factor of 10 if there is no change in the validation perplexity for a fixed number of epochs. BLEU~\cite{papineni:2002} and TER~\cite{snover:2006:ter} scores are computed with tokenized lower-cased output and references using the "evaluater" binary from Moses.

\begin{table}[!hbt]
	\centering
	\begin{tabular}{l|c|c}
		  & BLEU & 1-TER \\ \hline
		 baseline & 30.75 & 50.71 \\ \hline
		 no more split* & 30.74 & 49.82 \\ \hline
		 protect unseen,K=1 & 29.44 & 47.78 \\
		 protect unseen,K=2 & 30.15 & 48.78 \\
		 protect unseen,K=3 & 30.35 & 49.55 \\	
		 protect unseen,K=4 & 30.78 & 50.11 \\
		
	\end{tabular}
	\caption{Evaluation scores. K: threshold for the unk-count}
	\label{tab:results}
\end{table}

\begin{table*}[!hbt] \small
	\centering
	\begin{tabular}{|p{1.7cm}|p{12.5cm}|}
		\hline
		input & die idee hinter \textbf{littlebits} ist , dass es eine wachsende bibliothek ist . (de) \\ \hline
		reference & \textit{the idea behind \textbf{littlebits} is that its a growing library .} (en) \\ \hline
		baseline & die idee hinter \textbf{li@@ tt@@ leb@@ it@@ s} ist , dass es eine wachsen@@ de bibliothe@@ k ist .(de) \\ 
		& \textit{the idea behind \textbf{littlement} is that its a growing library .} (en) \\ \hline
		no more split & die idee hinter \textbf{li@@ tt@@ leb@@ its} ist , dass es eine wachsende bibliothek ist .(de) \\ 
		& \textit{the idea behind \textbf{littlebits} is that its a growing library .} (en) \\ \hline
	\end{tabular}
	\caption{Comparison of translation on a sentence from test corpus}
	\label{tab:example}
\end{table*}

The evaluation scores are detailed in Table~\ref{tab:results}. The quality scores have not improved using the proposed methods, but in manual evaluation, it was found that the model trained with "no more split" setting preserves better the named entities. This is depicted with an example in Table~\ref{tab:example}. The model with "protect unseen" with threshold value of 4 is slightly better than baseline, but in manual evaluation, we have seen that it is not better at translating the named entities compared to the baseline. 

\section{Domain Adaptation}
\label{sect:adapt}
As shown by \newcite{koehn-knowles-2017-six}, NMT is even more sensitive to the domain than phrase-based SMT. Translation quality drops abruptly when the source text is in a different domain to the training data. A standard technique to adapt a generic model to a specific domain is to continue the training with a small amount of in-domain parallel data. This technique, referred to as fine-tuning, is very effective. 

Our translation models are dynamically adapted to the source text context at each sentence, using fine-tuning but without knowing the source domain in advance. This adaptation is performed with a method similar to that proposed by \newcite{farajian-etal-2017-multi}. If a segment similar to the source sentence is found in the training corpus, the model is fine-tuned with the corresponding sentence pair for a few epochs. To this end, the training corpus is indexed into a translation memory. At test time, the translation memory is queried with the source sentence by information retrieval tools\footnote{Concretely we use Lucene~\cite{McCandless:2010:LAS:1893016}, a very efficient open-source information retrieval library.}. The number of epochs and the learning rate of the fine tuning with the retrieved sentence pair depends on the similarity between its source side and the source sentence. If they are not similar, fine tuning the model with the retrieved sentence may worsen the translation. The more they are similar, the more fine tuning can be beneficial and thus the higher the learning rate and number of epochs. This technique has thus more impact when the source text is very close to the training data.

We ran our pipeline with dynamic domain adaptation on the KDE4 German--English task (see Tables~\ref{tab:kde4:stats} and \ref{tab:cleaning:results}). The results are shown in Table~\ref{tab:adaptation:results}
 \begin{table}[!hbt]
	\centering
	\begin{tabular}{l|c|c}
		  & BLEU & 1-TER \\ \hline
		 without adaptation & 33.5 & 50.2 \\
		 with adaptation & 34.1 & 50.7
	\end{tabular}
	\caption{Evaluation scores for dynamic domain adaptation.}
	\label{tab:adaptation:results}
\end{table}

The impact of dynamic adaptation on this corpus is positive according to automated metrics, but modest. This is because for most sentences in the test set, there is no sentence in the translation memory being similar enough to fine-tune the model on it (see ~\newcite{farajian-etal-2017-multi} for more details).
Table~\ref{tab:adapt:example} shows an adaptation example. After fine-tuning on the corpus sentence pair "Gr\"{o}{\ss}e des Verlaufs@@ speichers :"--"clipboard history size :" (same as the source with a semicolon at the end), the model does not omit the word "Clipboard" any more.

\begin{table}[!hbt] \small
	\centering
	\begin{tabular}{|p{1.5cm}|p{5.2cm}|}
		\hline
		input & Gr\"{o}{\ss}e des Verlaufsspeichers \\ \hline
		reference & Clipboard history size \\ \hline
		baseline & History size \\ \hline
	    adapted	& Clipboard history size \\ \hline
		TM source & Gr\"{o}{\ss}e des Verlaufsspeichers : \\ \hline
		TM target & clipboard history size : \\ \hline
	\end{tabular}
	\caption{Example of dynamic adaptation.}
	\label{tab:adapt:example}
\end{table}



\section{What does all of this mean in practice?}
In real-world MT scenarios, it is often the finer details around the edges that can be of most importance. For example, in legal use cases like e-discovery, it is critical to get entities like names and addresses correct, because the resulting output is not being read by people, but rather being input into search tools where these entities will likely be search terms.

In other cases, such as MT for post-editing, where an end user will be working with the output, we may need the flexibility to act on specific feedback in order to address issues or concerns with the output.

The issues described above can manifest themselves in general, untrained engines, and the techniques we apply require an understanding of what is happening in the model, and the ability to be able to affect change. Then, finally, building upon strong baseline models to produce the most effective output for an particular use case. 

When looking at automated metrics, the impact of these techniques may not be very apparent, further emphasizing the need to human assessments prior to deploying an engine in production, particularly in certain scenarios.

\bibliography{mtsummit2019}
\bibliographystyle{mtsummit2019}

\end{document}